\documentclass{sig-alternate}

\usepackage[utf8]{inputenc}
\usepackage[table]{xcolor}
\begin{document}


\toappear{This article is an extended version of a poster article accepted at GECCO~2015, available under DOI: http://dx.doi.org/10.1145/2739482.2764678}
\clubpenalty=10000
\widowpenalty = 10000

\title{Model Selection and Overfitting in Genetic Programming: Empirical Study}
\subtitle{[Extended Version]}

%
%
%
%
%

\numberofauthors{2} 
%
\author{
%
%
\alignauthor
Jan Žegklitz\\
       \affaddr{Czech Technical University in Prague}\\
       \affaddr{Technická 2, Prague 6, Czech Republic}\\
       \email{zegkljan@fel.cvut.cz}
\alignauthor
Petr Pošík\\
       \affaddr{Czech Technical University in Prague}\\
       \affaddr{Technická 2, Prague 6, Czech Republic}\\
       \email{petr.posik@fel.cvut.cz}
}

\maketitle
\begin{abstract}
Genetic Programming has been very successful in solving a large area of
problems but its use as a machine learning algorithm has been limited so far.
One of the reasons is the problem of overfitting which cannot be solved or
suppresed as easily as in more traditional approaches. Another problem, closely
related to overfitting, is the selection of the final model from the population.

In this article we present our research that addresses both problems: overfitting
and model selection. We compare several ways of dealing with ovefitting, based
on Random Sampling Technique (RST) and on using a validation set, all with
an emphasis on model selection. We subject each approach to a thorough testing
on artificial and real--world datasets and compare them with the standard
approach, which uses the full training data, as a baseline.
\end{abstract}




\section{Introduction} \label{sec:introduction}

In machine learning (ML), the general task is to find a model that is able to
predict certain values (unknown in advance) of some objects based on a set of
known features of the particular object. There are two kinds of this task:
classification and regression. In classification the task is to assign a class
(from a finite set of classes) to a given object. In the most general case the
class is just a label and there is no other property that the set of classes
has (e.g. they need not be orderable). In regression the task is to assign a
quantitative value to a given object. In supervised ML, this is achieved using
a set of objects with known target class/value to train a particular model.

The need for a correct method to fit the model, tune its metaparameters, select
the final model, and estimate its error was recognized a long time ago. In the
first attempts, a model was fit to all the available data, and the error of the
model on this data was reported. However, as it was soon found out, such a
number often underestimated the true error observed on new unseen data, and
that such models are not very useful for prediction. This phenomenon -- small
error reported after training and large error observed on new data -- is called
\emph{overfitting} and is caused by the model being fit to the small deviations
in the data (e.g. noise) rather then the general trends.

To get an unbiased estimate of the prediction error, the standard practice is
to split the available data into two disjoint sets, training and testing, fit
the model parameters to the training set, and estimate the prediction error on
the testing set. This method is sufficient to get an unbiased estimate of the
prediction error for the model, which was constructed by a particular instance
of a particular fitting algorithm.

However, there is a need to compare models of various types (results of
various fitting algorithms), or models constructed by the same fitting algorithm
with different metaparameters. When the testing error estimate is used for
model selection, the information about the testing set leaks into the process
of model learning, of which the model selection is an unseparable part, and the
reported testing error of the final model underestimates the prediction error
again. A common basic technique is thus to split the available data into three
sets: training, validation, and testing, which serve to fit a particular model,
select among available trained models, and estimate the prediction error,
respectively.

Genetic Programming (GP) is an evolutionary technique designed to find
structured solutions, such as mathematical expressions or computer programs,
well fit for a partiuclar task. GP can be applied to ML tasks too, evolving
trees wich represent classification or regression models.

\subsection{GP as a model fitting algorithm}
When GP is used in ML to evolve a model, there are several important
differences when compared to ordinary fitting methods:
\begin{itemize}
    \item Ordinary fitting methods often optimize the model parameters only,
    not its structure, because they rely on the structure of the model. On the
    one hand, this limits the class of models that can be generated, on the
    other hand, one can take advantage of this and search classes of different
    complexities separately. GP, often even with fixed meta-parameters,
    produces highly free-form models, i.e. models with very different
    structures with a broad range of complexity.
    \item Ordinary fitting methods are often deterministic, gradient, or
    best-first search, algorithms, often quickly converging to a local optimum
    of the objective function. GP is basically a stochastic search, slow,
    without any guarantee that at least local optimum was found to certain
    precision.
    \item Ordinary fitting methods usually allow to fit the models with respect
    to a single training dataset; the error measured on this training data
    drives the parameter optimization process. Despite theoretically possible,
    they usually do not allow to store a separate best model so far with
    respect to a different data set. In GP, on the other hand, it is quite easy
    to use two or more datasets, one for driving the evolution process,
    and the other for best model selection.
\end{itemize}

GP as a ML method can work in two basic modes:
\begin{enumerate}
    \item GP can be treated as any other ML model fitting algorithm, i.e. we
    can fix/optimize its meta-parameters, like population size, the set of
    function symbols to be used, the maximal tree size
    and depth, etc. The best model in the sense of training error is provided
    as output. To select the best model, various combinations of
    meta-parameters can be tried, the resulting models can be evaluated on the
    validation set, and the best of them is chosen.

    Downsides: many algorithm runs, unstable results even for the same data,
    many meta-parameters value combinations to evaluate.
    \item GP can be run in a relatively non-limiting setting allowing it to
    create models of wide range of complexities, thus searching many model
    complexity classes at once. The GP algorithm generates many candidate
    models this way (all population members of all generations); in such a
    setting, however, the models are likely to overfit the training data and
    other means are needed to limit the influence of overfitting.
\end{enumerate}
In this article, we would like to concentrate on the second operation mode of
GP, comparing several means to limit overfitting, with the goal to asses the
influence of individual methods on final model results.

\subsection{Related work}
Bloat is a phenomenon in GP which can be described as an uncontrolled growth of
the program size a very small or no impact on the fitness. Several succesful
bloat control techniques were developed (e.g. \cite{poli:parsimony} and
\cite{silva:dynlimits}). The problem of overfitting was
often put into correlation with bloat. This was led by the ideas that bloated
models are more likely to be able to fit the noise rather than the short
models. However, it was shown \cite{vanneschi:opeq} that even in a bloat-free
envirnoment overfitting can still occur.

A technique called Random Subset Selection or Random Sampling Technique was
previously used for the speedup of the GP run \cite{gathercole:rss} and for
reducing overfitting \cite{liu:rst}. This technique was then further explored
in \cite{goncalves:vs, goncalves:rst}. These methods appeared to be successful
both in reducing the runtime and overfitting.

\section{Overfitting and Model Selection in GP}\label{sec:overfitting_selection}
In GP as a ML algorithm, there are two tasks the evolutionary algorithm must
perform:
\begin{enumerate}
    \item Drive the evolution, i.e. use such fitness that leads to better
    solutions.
    \item Be able to return a single ,,final`` model in any generation.
\end{enumerate}

In the rest of this section we are going to focus on methods of performing
these two tasks that are based solely on how the data are handled.

In all approaches we are going to tackle in the rest of the article the data
set is initially divided into two parts. The first part we call
\emph{training data} (TRN) and it is the data that are the input to the
particular GP algorithm. The second part we call \emph{testing data} (TST) and
it is the data used to evaluate the performance of the final model. The testing
data are never available to the algorithm during learning.

For all the methods presented in this article we can further divide the TRN
data to two subsets. The first one is used primarily to drive the evolution
(i.e. compute fitness) and we call this subset \emph{training data A} (TRN-A).
The other one is used primarily to select the model best so far  and we call
this subset \emph{training data B} (TRN-B). These two subsets can either be
disjoint or they can overlap or even be identical.

\subsection{Standard GP} \label{sec:standard-gp}
\begin{figure}[h]
    \centering
    \epsfig{file=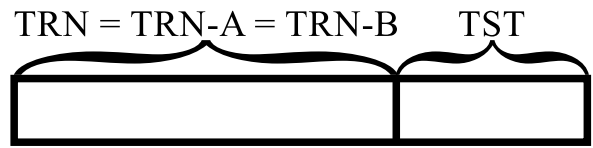}
    \caption{Illustration of data division in standard GP approach. TRN, TRN-A
    and TRN-B sets are all identical.}
    \label{fig:standard-gp-data}
\end{figure}

In the standard approach the whole TRN set is used for both tasks presented in
Section \ref{sec:overfitting_selection}, i.e. the fitness is the error on the
TRN set and the model selection is performed by storing the best model so far
with respect to the error TRN set too, i.e. fitness. From the perspective of
the data division, all the three sets TRN, TRN-A and TRN-B are identical. The
illustration of the data division can bee seen in the Figure
\ref{fig:standard-gp-data}.

\subsection{Validation set based approaches}
\begin{figure}[h]
    \centering
    \epsfig{file=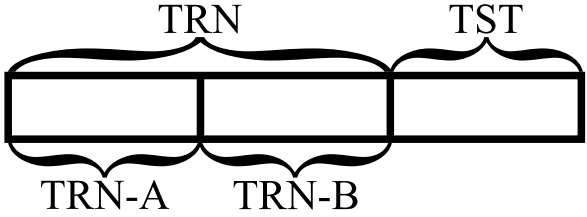}
    \caption{Illustration of data division in validation set based approaches.}
    \label{fig:validation-data}
\end{figure}

In the approaches based on validation set the TRN-A and TRN-B sets are
disjoint. In these approaches the TRN-B set aids in avoiding overfitting and
the model selection at the same time. The illustration of the data division can
be seen in the Figure \ref{fig:validation-data}.

For the purposes of this article we review two of those approaches: Backwarding
\cite{robilliard:backwarding} and Validation Start \cite{goncalves:vs}.

\subsubsection{Backwarding}
In Backwarding (BW) the evolution is driven only by the TRN-A set, i.e. the
fitness is the error over TRN-A. However, the model is selected according to
the TRN-B set by storing the best model so far with respect to the error on
this set.

This approach has the advantage that the model selection is not based on the
data used to learn the model itself and is therefore less likely to produce
overfitted models.

\subsubsection{Validation Start}
In Validation Start (VS) both the evolution and model selection is driven by
both sets TRN-A and TRN-B. The fitness is calculated as a weighted sum of two
components: error over TRN-A and the absolute difference of the error over
TRN-A and the error over TRN-B. The weights $w_1$ and $w_2$ are both random but
correlated such that $w_1 + w_2 = 1$; they stay fixed for the whole run of the
algorithm.

The model selection is done according to the fitness, i.e. the final model is
the one with the best fitness.

This approach is motivated by the fact that we want the models to have similar
error both on the ,,true`` training data (TRN-A in our terminology) and on the
data not used for the actual training.

\subsection{Random Sampling based approaches}
Random Sampling Technique (RST) \cite{goncalves:vs, goncalves:rst} is based on
the idea of using only a random subset of the training data for fitness
evaluation, changing this subset during the evolution.

The model selection is done with respect to the error over the whole TRN set,
i.e. TRN-B is identical to TRN.

\subsubsection{RST 1/1}
In \cite{goncalves:vs} it was shown that the extreme case of using only a
single-element subset and changing it every generation, called RST 1/1,
produced the best results with respect to the testing error and hence acts as a
good technique to control overfitting.

\begin{figure}[h]
    \centering
    \epsfig{file=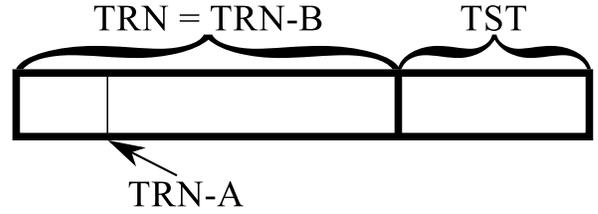}
    \caption{Illustration of data division in RST 1/1. The TRN-A set is a
    single element changing every generation.}
    \label{fig:rst-single-data}
\end{figure}

In our terminology of data division, the TRN-A set is composed of a single
element chosen randomly from the TRN set and changes every generation. The
TRN-B set is identical to the TRN set. The illustration of the data division
can be seen in the Figure \ref{fig:rst-single-data}.

\subsubsection{Random Interleaved}
Random Interleaved (RI) \cite{goncalves:ri} is a technique based on RST~1/1. It
is motivated by the fact that in RST only a very small fraction of information
is used to learn the model.

In RI, each generation, one of two possibilities of fitness evaluation
is randomly chosen. One possibility is identical to RST 1/1, i.e. the fitness
is evaluated on a single data point, and the other possibility is identical
to standard approach, i.e. the fitness is evaluated on the whole TRN set.
The RI method is parametrised by the percentage $P \%$ which indicates the
probability of choosing RST 1/1 as the evaluation method in a generation.

\begin{figure}[h]
    \centering
    \epsfig{file=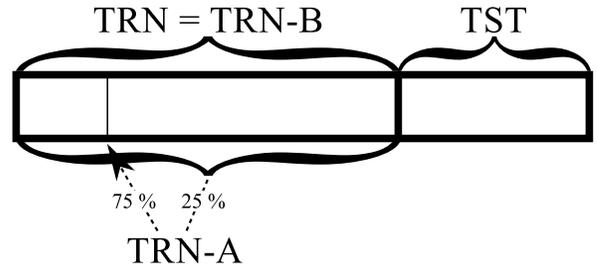}
    \caption{Illustration of data division in RI 75\%.}
    \label{fig:ri-data}
\end{figure}

In our terminology, in $P \%$ of generations (on average) the TRN-A set is of a
single element randomly chosen from TRN and TRN-B is identical to TRN, and in
$100 - P \%$ of generations the TRN, TRN-A and TRN-B sets are all identical.
The illustration of the data division (for the example of RI 75 \%) can be seen
in the Figure \ref{fig:ri-data}.

\subsubsection{RST R}
Inspired by the succes of RI 50 \% \cite{goncalves:ri} we propose a variant of
RST which follows the same motivation as RI but achieves it in a different way.
We call our variant RST R (the R stands for ,,random``). It is almost identical
to RST but not only the elements of the subset are chosen randomly, the size of
this subset is chosen randomly too. The size is drawn from a uniform distribution
resulting in using (on average) 50 \% of the data points (repetitions counted),
almost as in RI 50 \%.

\subsection{Combining Random Sampling with validation set}
All the techniques based on Random Sampling effectively use the whole TRN set
both for driving the evolution, which can be seen as the training itself, and
for selecting the best model. As we already mentioned in the Section
\ref{sec:introduction}, from the point of view of traditional ML methods, this
approach is not desirable and can lead to models selected based on
underestimated errors.

To improve the algorithms in this area we propose to combine validation set
approach with the Random Sampling approach.
\\
\subsubsection{VRST 1/1}
VRST 1/1 stands for validation-RST 1/1 and is a variant of RST 1/1 with validation
set.

\begin{figure}[h]
    \centering
    \epsfig{file=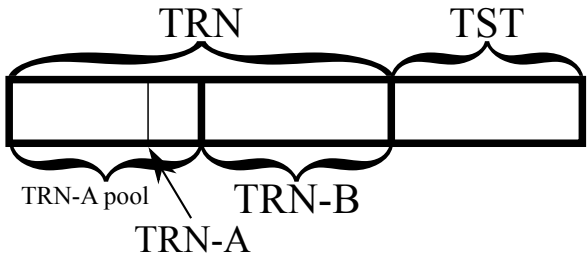}
    \caption{Illustration of data division in VRST 1/1. The TRN-A is a single
    element drawn from the TRN-A pool.}
    \label{fig:vrst-data}
\end{figure}

In VRST 1/1, the TRN-A set is a single element changed every generation, like in
RST 1/1, but it is drawn from a ,,TRN-A pool`` which is disjoint from the TRN-B
set. The best model so far is determined using the TRN-B set. The illustration of
data division can be seen in the Figure \ref{fig:vrst-data}.

\subsubsection{VRI}
VRI stands for validation-RI and is a variant of RI with validation set.

\begin{figure}[h]
    \centering
    \epsfig{file=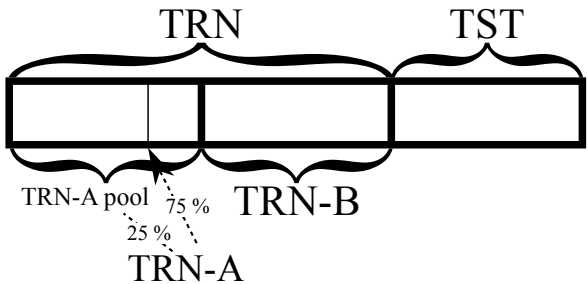}
    \caption{Illustration of data division in VRI 75\%. The 75\% case of TRN-A is
    drawn from the TRN-A pool.}
    \label{fig:vri-data}
\end{figure}

In VRI, the TRN-A set is either a single element sampled from a ,,TRN-A pool`` or
the whole pool (depends on the percentage parameter). The TRN-A pool and TRN-B sets
are disjoint. The best model so far is determined using the TRN-B set. The
illustration of data division (for the example of VRI 75\%) can be seen in the
Figure \ref{fig:vri-data}.

\subsubsection{VRST R}
VRST R stands for validation-RST R and is a variant of RST R with validation set.

In VRST R, the TRN-A set is a subset of ,,TRN-A pool`` of random (uniformly distributed)
size. The TRN-A pool and TRN-B sets are disjoint. The best model so far is determined
using the TRN-B set.

\section{Experimental evaluation}
In the previous section we reviewed or proposed various approaches to prevent
overfitting. In order to find out how each approach works, we conducted a
series of experiments on various datasets. The rest of this section describes
the used datasets and the setup of the algorithms and experiments.

\subsection{Data sets} \label{sec:datasets}
For the testing of all the algorithms we used a set of six datasets.

Two Spirals (TS), Cluster in Cluster (CIC) and Halfkernel (HK) are binary
classification datasets generated by MATLAB scripts from \cite{matlab:datasets}.

Sphere (SPH) is a regression dataset defined as
\begin{displaymath}
    f(\mathbf{x}) = \sum_{i=0}^{30} x_i^2 + noise
\end{displaymath}
where each $x_i$ is independently randomly sampled from interval
$\left[-1.5, 1.5\right]$ and $noise$ is a random value drawn uniformly from
interval $\left[-6, 6\right]$.

Forest Fires (FF) \cite{uci:forestfires} retrieved from the UCI repository
\cite{uci} is a real-world regression dataset where the task is to predict the
burned area of the forest. All features are numeric except the 3rd and 4th
features which are month (,,jan`` to ,,dec``) and day (,,mon`` to ,,sun``)
respectively. These were transformed to numbers by mapping the month to the
numbers 1 to 12 (,,jan`` being mapped to 1, ,,dec`` being mapped to 12) and day
to the numbers 1 to 7 (,,mon`` being mapped to 1, ,,sun`` being mapped to 7).

Wisconsin Diagnostic Breast Cancer (WDBC) retrieved from the UCI repository
\cite{uci} is a real-world binary classification dataset where the task is to
state a diagnosis (malignant/benign) based on numeric features computed from
digitized image of a breast mass.

A summary description of the datasets is in Table \ref{tab:datasets}.
\begin{table}
    \centering
    \begin{tabular}{|c|c|c|c|c|}
        \hline
        dataset & artificial & task  & \# instances & \# features \\ \hline
        TS      &        yes & clas. & 3000         & 2 \\ \hline
        CIC     &        yes & clas. & 1240         & 2 \\ \hline
        HK      &        yes & clas. & 1200         & 2 \\ \hline
        SPH     &        yes & regr. & 1500         & 30 \\ \hline
        FF      &         no & regr. & 517          & 12 \\ \hline
        WDBC    &         no & clas. & 569          & 30 \\ \hline
    \end{tabular}
    \caption{Summary information about the used datasets.}
    \label{tab:datasets}
\end{table}
All classification tasks are binary ones (i.e. there are two classes to
classify).

\subsection{GP algorithm and setup}
For the GP algorihm we used Grammatical Evolution \cite{ge}. The setup of the algorithm
was identical in all the used methods and datasets and can be seen in the Table
\ref{tab:ge-setup}.
\begin{table}[ht]
    \centering
    \begin{tabular}{|c|c|}
    \hline
    initial maximum genotype length & 100 codons \\
    \hline
    codon value range & 0 to 255 \\
    \hline
    pop. size & 500 \\
    \hline
    generations & 200 \\
    \hline
    selection & tournament of 4 \\
    \hline
    crossover prob. & 0.5 \\
    \hline
    crossover type & single-point (ripple) \\
    \hline
    mutation prob. (per codon) & 0.1 \\
    \hline
    mutation type & \begin{minipage}[t]{0.35\columnwidth}\centering change codon to random number from range\end{minipage} \\
    \hline
    pruning prob. & 0.2 \\
    \hline
    duplication prob. & 0.2 \\
    \hline
    maximum wraps & 0 (wrapping disabled) \\
    \hline
    \end{tabular}
    \caption{Setup of the GE algorithm.}
    \label{tab:ge-setup}
\end{table}

For all the experiments we used the same grammar which is described in the Appendix \ref{sec:grammar}.

For regression tasks, the output of the evolved expression was directly used as
the estimated value. For classification tasks (only binary classification, see
Section \ref{sec:datasets}), if the output of the evolved expression was less
than 0 then the first class was assigned, else was the second class.

The (V)RI methods' percentage paremeter was set to 60\% as it is a middle ground
between 50\% and 75\% that were among the most successful ones in
\cite{goncalves:ri}.

If any solution produced a mathematical error (e.g. division by zero or a logarithm
of negative number) it received infinite fitness (which is always the worst).
\\
\\

\subsection{Setup of the experiments}
For each dataset every algorithm was run 96 times. The TRN/TST division ratio
was fixed to 70\%/30\% for all runs of all algorithms on all datasets.
The division was different for each run of an algorithm but runs with equal
number had the same division (e.g. the first runs of RST 1/1 and VRST R on TS
dataset had the same TRN and TST subsets).

Algorithms BW, VS, VRI, VRST R and VRST 1/1 had the TRN-A/TRN-B division ratio
fixed to 50\%/50\%, again different in every run but equal in corresponding
runs of all the algorithms.

\section{Results} \label{sec:results}
\begin{figure}
    \centering
    \epsfig{file=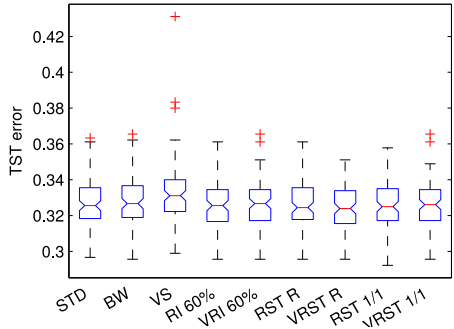, width=6cm, bbllx=.1cm, bblly=.3cm, bburx=9.9cm, bbury=7.5cm, clip=}
    \caption{Box plot of the final values of the TST error on the TS (Two
    Spirals) dataset.}
    \label{fig:ts-tst-box}
\end{figure}

\begin{figure}
    \centering
    \epsfig{file=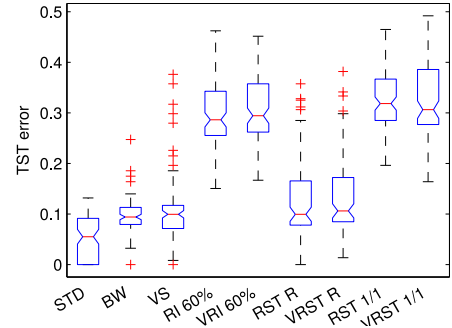, width=6cm, bbllx=.1cm, bblly=.3cm, bburx=9.9cm, bbury=7.5cm, clip=}
    \caption{Box plot of the final values of the TST error on the CIC (Cluster
    In Cluster) dataset.}
    \label{fig:cic-tst-box}
\end{figure}

\begin{figure}
    \centering
    \epsfig{file=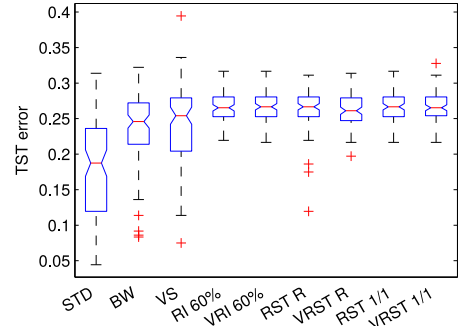, width=6cm, bbllx=.1cm, bblly=.3cm, bburx=9.9cm, bbury=7.5cm, clip=}
    \caption{Box plot of the final values of the TST error on the HK
    (Halfkernel) dataset.}
    \label{fig:hk-tst-box}
\end{figure}

\begin{figure}
    \centering
    \epsfig{file=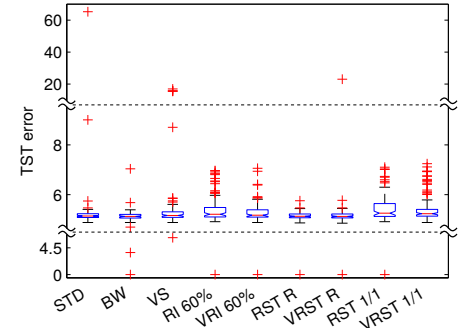, width=6cm, bbllx=.1cm, bblly=.3cm, bburx=9.9cm, bbury=7.5cm, clip=}
    \caption{Box plot of the final values of the TST error on the SPH (Sphere)
    dataset.}
    \label{fig:sph-tst-box}
\end{figure}

\begin{figure}
    \centering
    \epsfig{file=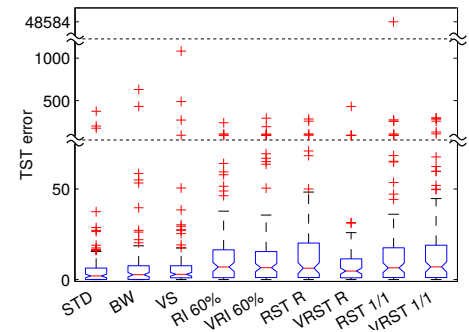, width=6cm, bbllx=.1cm, bblly=.3cm, bburx=9.9cm, bbury=7.5cm, clip=}
    \caption{Box plot of the final values of the TST error on the FF (Forest
    Fires) dataset.}
    \label{fig:ff-tst-box}
\end{figure}

\begin{figure}
    \centering
    \epsfig{file=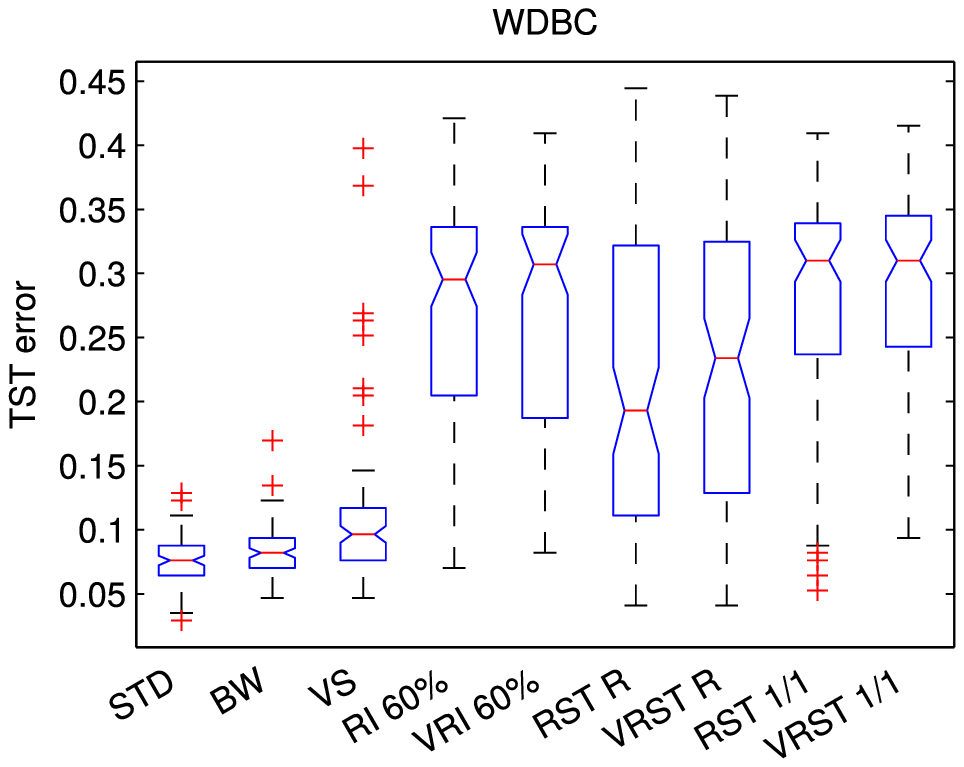, width=6cm, bbllx=.1cm, bblly=.3cm, bburx=9.9cm, bbury=7.5cm, clip=}
    \caption{Box plot of the final values of the TST error on the WDBC
    (Wisconsin Diagnostic Breast Cancer) dataset.}
    \label{fig:wdbc-tst-box}
\end{figure}

Box plots of the final values of TST error can be seen in~Figures
\ref{fig:ts-tst-box}, \ref{fig:cic-tst-box}, \ref{fig:hk-tst-box},
\ref{fig:sph-tst-box}, \ref{fig:ff-tst-box} and \ref{fig:wdbc-tst-box}.

The overall results can be seen in the Table \ref{tab:overall-results}. The
,,rank`` column in this table is the rank of the method on the particular
dataset. The rank was determined using the one-sided Mann-Whitney U test on the
final TST errors, pairwise among all methods on the particular dataset, with the
level of significance $\alpha = 0.05$: methods with equal ranks are not
statistically significantly different; if two methods, A and B, have ranks $r_A$
and $r_B$ such that $r_A < r_B$ then method A is statistically significantly
better than method~B.
\begin{table*}
    \centering
    \begin{tabular}{|c|c|c|r|r|r||r|r|r||r|r|r|}
\cline{4-12}
\multicolumn{3}{c|}{}     &                                  \multicolumn{3}{c||}{TST error} &                                   \multicolumn{3}{c|}{TRN error} & \multicolumn{3}{c||}{tree size} \\ \hline
dataset & rank & method   &                    median &                        mean & stddev &                    median &                       mean &  stddev &   median &   mean & stddev \\ \hline\hline
TS      &    1 & STD      & \cellcolor{gray!14} 0.326 & \cellcolor{gray!16}   0.326 &  0.014 & \cellcolor{gray!00} 0.321 & \cellcolor{gray!00}  0.321 &   0.006 &        5 &  6.844 &  4.913 \\ \cline{3-12}
        &      & BW       & \cellcolor{gray!23} 0.327 & \cellcolor{gray!23}   0.328 &  0.014 & \cellcolor{gray!44} 0.325 & \cellcolor{gray!21}  0.324 &   0.006 &        6 &  6.677 &  3.862 \\ \cline{3-12}
        &      & RST 1/1  & \cellcolor{gray!09} 0.325 & \cellcolor{gray!07}   0.325 &  0.013 & \cellcolor{gray!35} 0.324 & \cellcolor{gray!19}  0.324 &   0.006 &        3 &  3.604 &  2.255 \\ \cline{3-12}
        &      & VRST 1/1 & \cellcolor{gray!18} 0.326 & \cellcolor{gray!10}   0.326 &  0.014 & \cellcolor{gray!44} 0.325 & \cellcolor{gray!25}  0.325 &   0.005 &        3 &  3.365 &  2.304 \\ \cline{3-12}
        &      & RST R    & \cellcolor{gray!05} 0.324 & \cellcolor{gray!09}   0.325 &  0.013 & \cellcolor{gray!33} 0.324 & \cellcolor{gray!14}  0.323 &   0.006 &        3 &  4.229 &  3.197 \\ \cline{3-12}
        &      & VRST R   & \cellcolor{gray!00} 0.324 & \cellcolor{gray!00}   0.324 &  0.013 & \cellcolor{gray!44} 0.325 & \cellcolor{gray!23}  0.324 &   0.006 &        3 &  4.458 &  5.013 \\ \cline{3-12}
        &      & RI 60\%  & \cellcolor{gray!14} 0.326 & \cellcolor{gray!09}   0.325 &  0.013 & \cellcolor{gray!38} 0.324 & \cellcolor{gray!20}  0.324 &   0.006 &        3 &  3.740 &  3.079 \\ \cline{3-12}
        &      & VRI 60\% & \cellcolor{gray!23} 0.327 & \cellcolor{gray!11}   0.326 &  0.013 & \cellcolor{gray!44} 0.325 & \cellcolor{gray!25}  0.325 &   0.005 &        3 &  3.396 &  2.693 \\ \cline{2-12}
        &    2 & VS       & \cellcolor{gray!60} 0.331 & \cellcolor{gray!60}   0.333 &  0.018 & \cellcolor{gray!60} 0.326 & \cellcolor{gray!60}  0.330 &   0.018 &        8 &  8.135 &  5.364 \\ \hline\hline
CIC     &    1 & STD      & \cellcolor{gray!00} 0.055 & \cellcolor{gray!00}   0.054 &  0.044 & \cellcolor{gray!00} 0.054 & \cellcolor{gray!00}  0.051 &   0.042 &       11 & 11.500 &  3.393 \\ \cline{2-12}
        &    2 & BW       & \cellcolor{gray!09} 0.094 & \cellcolor{gray!09}   0.093 &  0.038 & \cellcolor{gray!10} 0.095 & \cellcolor{gray!08}  0.089 &   0.032 &       10 & 10.427 &  2.944 \\ \cline{3-12}
        &      & VS       & \cellcolor{gray!10} 0.099 & \cellcolor{gray!12}   0.107 &  0.069 & \cellcolor{gray!10} 0.094 & \cellcolor{gray!12}  0.104 &   0.069 &       10 & 10.542 &  3.346 \\ \cline{3-12}
        &      & RST R    & \cellcolor{gray!10} 0.099 & \cellcolor{gray!15}   0.121 &  0.074 & \cellcolor{gray!11} 0.099 & \cellcolor{gray!14}  0.117 &   0.069 &        9 &  9.271 &  3.144 \\ \cline{3-12}
        &      & VRST R   & \cellcolor{gray!12} 0.106 & \cellcolor{gray!17}   0.132 &  0.071 & \cellcolor{gray!13} 0.107 & \cellcolor{gray!18}  0.134 &   0.068 &        8 &  8.823 &  3.369 \\ \cline{2-12}
        &    3 & RI 60\%  & \cellcolor{gray!53} 0.286 & \cellcolor{gray!52}   0.292 &  0.068 & \cellcolor{gray!56} 0.291 & \cellcolor{gray!53}  0.292 &   0.059 &        6 &  6.854 &  4.282 \\ \cline{2-12}
        &    4 & RST 1/1  & \cellcolor{gray!60} 0.319 & \cellcolor{gray!60}   0.329 &  0.060 & \cellcolor{gray!60} 0.308 & \cellcolor{gray!60}  0.324 &   0.050 &        4 &  5.552 &  3.825 \\ \cline{3-12}
        &      & VRST 1/1 & \cellcolor{gray!57} 0.306 & \cellcolor{gray!59}   0.326 &  0.065 & \cellcolor{gray!58} 0.301 & \cellcolor{gray!59}  0.321 &   0.055 &        4 &  5.396 &  2.871 \\ \cline{3-12}
        &      & VRI 60\% & \cellcolor{gray!54} 0.294 & \cellcolor{gray!55}   0.308 &  0.070 & \cellcolor{gray!56} 0.291 & \cellcolor{gray!56}  0.304 &   0.060 &        4 &  5.292 &  2.768 \\ \hline\hline
HK      &    1 & STD      & \cellcolor{gray!00} 0.188 & \cellcolor{gray!00}   0.178 &  0.066 & \cellcolor{gray!00} 0.175 & \cellcolor{gray!00}  0.174 &   0.059 &       12 & 12.417 &  4.106 \\ \cline{2-12}
        &    2 & BW       & \cellcolor{gray!44} 0.246 & \cellcolor{gray!38}   0.235 &  0.050 & \cellcolor{gray!44} 0.244 & \cellcolor{gray!36}  0.231 &   0.042 &        9 & 10.281 &  4.436 \\ \cline{3-12}
        &      & VS       & \cellcolor{gray!51} 0.254 & \cellcolor{gray!45}   0.244 &  0.054 & \cellcolor{gray!47} 0.248 & \cellcolor{gray!43}  0.241 &   0.050 &        9 &  9.812 &  4.409 \\ \cline{2-12}
        &    3 & RST 1/1  & \cellcolor{gray!60} 0.267 & \cellcolor{gray!60}   0.266 &  0.021 & \cellcolor{gray!59} 0.267 & \cellcolor{gray!59}  0.267 &   0.009 &        3 &  3.833 &  2.638 \\ \cline{3-12}
        &      & VRST 1/1 & \cellcolor{gray!59} 0.265 & \cellcolor{gray!60}   0.267 &  0.021 & \cellcolor{gray!60} 0.268 & \cellcolor{gray!60}  0.269 &   0.011 &        3 &  4.240 &  4.301 \\ \cline{3-12}
        &      & RST R    & \cellcolor{gray!60} 0.267 & \cellcolor{gray!57}   0.263 &  0.028 & \cellcolor{gray!56} 0.262 & \cellcolor{gray!53}  0.258 &   0.022 &        6 &  6.792 &  4.410 \\ \cline{3-12}
        &      & VRST R   & \cellcolor{gray!56} 0.261 & \cellcolor{gray!58}   0.263 &  0.021 & \cellcolor{gray!58} 0.265 & \cellcolor{gray!58}  0.265 &   0.011 &        4 &  5.385 &  3.310 \\ \cline{3-12}
        &      & RI 60\%  & \cellcolor{gray!59} 0.265 & \cellcolor{gray!60}   0.266 &  0.021 & \cellcolor{gray!59} 0.267 & \cellcolor{gray!59}  0.267 &   0.011 &        3 &  3.969 &  2.412 \\ \cline{3-12}
        &      & VRI 60\% & \cellcolor{gray!60} 0.267 & \cellcolor{gray!59}   0.266 &  0.020 & \cellcolor{gray!60} 0.268 & \cellcolor{gray!60}  0.268 &   0.011 &        3 &  4.010 &  2.606 \\ \hline\hline
SPH     &    1 & STD      & \cellcolor{gray!08} 5.141 & \cellcolor{gray!60}   5.821 &  6.144 & \cellcolor{gray!00} 5.127 & \cellcolor{gray!00}  5.069 &   0.529 &        8 &  9.646 &  3.995 \\ \cline{3-12}
        &      & BW       & \cellcolor{gray!00} 5.125 & \cellcolor{gray!00}   5.122 &  0.399 & \cellcolor{gray!02} 5.128 & \cellcolor{gray!05}  5.113 &   0.279 &        8 &  9.354 &  3.305 \\ \cline{3-12}
        &      & VS       & \cellcolor{gray!18} 5.163 & \cellcolor{gray!47}   5.667 &  2.202 & \cellcolor{gray!24} 5.152 & \cellcolor{gray!60}  5.601 &   2.130 &        9 &  9.854 &  3.043 \\ \cline{3-12}
        &      & RST R    & \cellcolor{gray!04} 5.132 & \cellcolor{gray!02}   5.150 &  0.137 & \cellcolor{gray!00} 5.126 & \cellcolor{gray!06}  5.122 &   0.169 &        8 &  9.000 &  2.220 \\ \cline{3-12}
        &      & VRST R   & \cellcolor{gray!00} 5.125 & \cellcolor{gray!18}   5.330 &  1.841 & \cellcolor{gray!03} 5.129 & \cellcolor{gray!16}  5.213 &   0.829 &        8 &  8.833 &  2.260 \\ \cline{3-12}
        &      & RI 60\%  & \cellcolor{gray!45} 5.222 & \cellcolor{gray!27}   5.434 &  0.517 & \cellcolor{gray!39} 5.168 & \cellcolor{gray!37}  5.400 &   0.496 &        8 &  8.167 &  2.360 \\ \cline{3-12}
        &      & VRI 60\% & \cellcolor{gray!22} 5.172 & \cellcolor{gray!17}   5.321 &  0.388 & \cellcolor{gray!54} 5.185 & \cellcolor{gray!28}  5.321 &   0.387 &        8 &  8.062 &  2.112 \\ \cline{2-12}
        &    2 & RST 1/1  & \cellcolor{gray!60} 5.253 & \cellcolor{gray!28}   5.450 &  0.507 & \cellcolor{gray!60} 5.191 & \cellcolor{gray!42}  5.444 &   0.509 &        8 &  8.167 &  2.482 \\ \cline{3-12}
        &      & VRST 1/1 & \cellcolor{gray!49} 5.229 & \cellcolor{gray!29}   5.462 &  0.582 & \cellcolor{gray!50} 5.180 & \cellcolor{gray!44}  5.462 &   0.574 &        8 &  8.302 &  1.985 \\ \hline\hline
FF      &    1 & STD      & \cellcolor{gray!00} 2.134 & \cellcolor{gray!00}  13.529 & 47.349 & \cellcolor{gray!00} 0.000 & \cellcolor{gray!00}  0.103 &   0.274 &        7 &  8.135 &  3.732 \\ \cline{3-12}
        &      & BW       & \cellcolor{gray!06} 2.720 & \cellcolor{gray!01}  18.004 & 78.684 & \cellcolor{gray!00} 0.000 & \cellcolor{gray!04}  4.599 &  31.233 &        7 &  9.010 &  6.141 \\ \cline{3-12}
        &      & VS       & \cellcolor{gray!10} 3.058 & \cellcolor{gray!02}  27.150 & 24.220 & \cellcolor{gray!00} 0.000 & \cellcolor{gray!00}  0.541 &   1.463 &        7 &  8.417 &  3.749 \\ \cline{3-12}
        &      & VRST R   & \cellcolor{gray!28} 4.813 & \cellcolor{gray!00}  15.634 & 48.126 & \cellcolor{gray!32} 4.859 & \cellcolor{gray!60} 73.167 & 540.151 &        7 &  8.354 &  6.614 \\ \cline{2-12}
        &    2 & RST 1/1  & \cellcolor{gray!52} 6.980 & \cellcolor{gray!60} 545.889 & 35.743 & \cellcolor{gray!33} 5.082 & \cellcolor{gray!10} 12.235 &  16.959 &        5 &  7.375 &  6.460 \\ \cline{3-12}
        &      & VRST 1/1 & \cellcolor{gray!60} 7.777 & \cellcolor{gray!02}  27.982 & 59.345 & \cellcolor{gray!60} 9.201 & \cellcolor{gray!11} 13.691 &  22.369 &        5 &  7.104 &  5.717 \\ \cline{3-12}
        &      & RST R    & \cellcolor{gray!45} 6.376 & \cellcolor{gray!01}  22.297 & 45.114 & \cellcolor{gray!07} 1.000 & \cellcolor{gray!07}  8.708 &  14.632 &        7 &  7.562 &  4.091 \\ \cline{3-12}
        &      & RI 60\%  & \cellcolor{gray!52} 6.986 & \cellcolor{gray!00}  17.076 & 31.035 & \cellcolor{gray!39} 5.936 & \cellcolor{gray!09} 11.064 &  13.201 &        3 &  6.344 &  4.847 \\ \cline{3-12}
        &      & VRI 60\% & \cellcolor{gray!51} 6.967 & \cellcolor{gray!01}  20.687 & 41.611 & \cellcolor{gray!40} 6.086 & \cellcolor{gray!09} 11.521 &  20.178 &        5 &  7.250 &  6.259 \\ \hline\hline
WDBC    &    1 & STD      & \cellcolor{gray!00} 0.076 & \cellcolor{gray!00}   0.078 &  0.020 & \cellcolor{gray!00} 0.065 & \cellcolor{gray!00}  0.066 &   0.012 &        8 &  8.083 &  3.201 \\ \cline{3-12}
        &      & BW       & \cellcolor{gray!02} 0.082 & \cellcolor{gray!01}   0.082 &  0.021 & \cellcolor{gray!02} 0.075 & \cellcolor{gray!02}  0.075 &   0.013 &        6 &  6.958 &  3.054 \\ \cline{2-12}
        &    2 & VS       & \cellcolor{gray!05} 0.096 & \cellcolor{gray!09}   0.109 &  0.058 & \cellcolor{gray!03} 0.080 & \cellcolor{gray!09}  0.100 &   0.059 &        6 &  7.167 &  4.410 \\ \cline{2-12}
        &    3 & RST R    & \cellcolor{gray!30} 0.193 & \cellcolor{gray!39}   0.215 &  0.111 & \cellcolor{gray!28} 0.187 & \cellcolor{gray!38}  0.209 &   0.106 &        6 &  6.729 &  4.826 \\ \cline{3-12}
        &      & VRST R   & \cellcolor{gray!41} 0.234 & \cellcolor{gray!43}   0.230 &  0.110 & \cellcolor{gray!38} 0.230 & \cellcolor{gray!45}  0.235 &   0.105 &        5 &  5.656 &  4.154 \\ \cline{2-12}
        &    4 & RST 1/1  & \cellcolor{gray!60} 0.310 & \cellcolor{gray!57}   0.279 &  0.086 & \cellcolor{gray!58} 0.318 & \cellcolor{gray!57}  0.279 &   0.079 &        3 &  4.792 &  5.651 \\ \cline{3-12}
        &      & VRST 1/1 & \cellcolor{gray!60} 0.310 & \cellcolor{gray!60}   0.288 &  0.078 & \cellcolor{gray!60} 0.327 & \cellcolor{gray!60}  0.290 &   0.074 &        3 &  4.260 &  4.223 \\ \cline{3-12}
        &      & RI 60\%  & \cellcolor{gray!56} 0.295 & \cellcolor{gray!56}   0.273 &  0.086 & \cellcolor{gray!57} 0.313 & \cellcolor{gray!54}  0.269 &   0.086 &        3 &  5.490 &  5.018 \\ \cline{3-12}
        &      & VRI 60\% & \cellcolor{gray!59} 0.307 & \cellcolor{gray!55}   0.270 &  0.090 & \cellcolor{gray!55} 0.304 & \cellcolor{gray!53}  0.265 &   0.085 &        3 &  5.156 &  5.135 \\ \hline
    \end{tabular}
    \caption{Table of results of the tested methods on all datasets. The
    darkness of the cell background is proportional to the values in the column
    for the respective dataset.}
    \label{tab:overall-results}
\end{table*}

\subsection{Performance of standard approach}
Surprisingly, on all the datasets, the standard approach was either the best
or in the group of the best approaches. This result is contradictory to
\cite{goncalves:vs, goncalves:rst} and \cite{goncalves:ri}. However, in
\cite{goncalves:vs} the approaches were tested only on a single one-dimensional
artificial dataset, and in \cite{goncalves:rst, goncalves:ri} the approaches
were tested on three real-world high-dimensional datasets. In our experiments
both artificial and real-world datasets of 2 to 30 dimensions were used.

This result suggests that the benefit of RST-based approaches is data-dependent
and general conclusion cannot be made based on the experiments carried out so
far.

\subsection{Benefit of using a validation set}
The only case where validation set variant of a method was statistically
significantly better than its non-validation set variant was the VRST R on FF
dataset. In all other cases the validation and non-validation variants of the
algorithms were not statistically significantly different.

This result suggests that using a validation set does not bring much benefit, at
least on the datasets and with setup used in these experiments. One of the
reasons might be the tradeoff between overfitting prevention and giving the
algorithm enough information to be able to learn.

\subsection{Random-sized subsets}
In all the experiments there was no case of the (V)RST~R being statistically
significantly worse than (V)RST~1/1 or (V)RI~60\%. On the other hand, on CIC and
WDBC datasets the (V)RST~R was statistically significantly better than both
(V)RI~60\% and (V)RST~1/1 and on FF dataset VRST R was statistically
significantly better than all other RST-based approaches.

This result suggests that using random-sized subsets might be more beneficial
than using either only a single-element subsets or switching between the full
set and single-element subset. The reason for this might be that when using a
single-element subset the number of fitness cases (meaning the part of the data
the solutions are to classify/model) the algorithm can encounter is much smaller
than in the case of random-sized subsets, causing lower variability of the
actual training data.

\section{Conclusions}
In this article we further explored the issue of overfitting in Genetic
Programming. In detail we discussed the ways the data are handled and based on
two patterns - validation set and random sampling - we proposed two new
approaches: RST with random-sized subsets and using a validation set in
RST-based techniques, including the combination of both.

The RST random was based on the idea that using only single element or only the
full training set makes the number of fitness cases small. RST R uses not only
randomly chosen elements of the subsets but the size of the subsets is random
too.

Bringing validation set to RST-based techniques was based on the idea that
selecting the model with respect to the actual training set could bring unwanted
bias towards the training data. In the variants with validation set the model
selection and the actual learning are isolated, preventing such bias.

We have carried out a series of experiments with all the presented approaches
on six datasets, both artificial and real-world, both for classification and
regression.

The most important result is that the standard approach, i.e. learn and select
model on the whole training set all the time, came out as either the best or
among the best approaches on all the datasets. This result is contradictory to
the previous articles \cite{goncalves:vs, goncalves:rst, goncalves:ri} where
this approach performed poorly. This indicates that the technique performance
could be highly data dependent and therefore we think no general conclusion
about the benefit of random sampling can be made. This result asks for further
investigation to find out which aspects of the data cause various approaches to
perform well or poorly.

The second result is that our idea of using a validation set did not prove to be
significantly beneficial. There was only a single case where the validation set
variant significantly outperformed the non-validation variant. However, this
approach also requires further investigation because it could also be data
dependent and also because we tested only one division ratio. Different setup
could have (or not) a significant impact on the performance of such methods.

The third result is the good performance of random-sized subsets with respect
to the other two RST-based methods. There was no case the random-sized subsets
caused worse performance than the other two methods and in some cases there
was a significant difference to the favor of the random-sized subsets. However,
this could be data dependent too, and further investigation is also needed.
Another aspect of this method is the distribution of the subset size. We used
uniform distribution but other distributions, e.g. favoring smaller subsets,
could prove even more beneficial.

\section{Acknowledgments}
This work was supported by the Grant Agency of the Czech Technical University
in Prague, grant No.\\ SGS14/194/OHK3/3T/13.

Access to computing and storage facilities owned by parties and projects
contributing to the National Grid Infrastructure MetaCentrum, provided under
the programme\\"Projects of Large Infrastructure for Research, Development, and
Innovations" (LM2010005), is greatly appreciated.

%
\bibliographystyle{abbrv}
\bibliography{bib}  
%
%
\balancecolumns
\appendix
\section{Grammar} \label{sec:grammar}
The grammar used for all experiments was of the following form (the italic texts are comments, not part of the grammar):

\noindent
\begin{tabular}{rrll}
    \texttt{<expr>}  & \texttt{::=} & \multicolumn{2}{l}{\texttt{(<expr> <biop> <expr>)}} \\
                     &   \texttt{|} & \multicolumn{2}{l}{\texttt{(<unop> <expr>)}} \\
                     &   \texttt{|} & \multicolumn{2}{l}{\texttt{<var>}} \\
                     &   \texttt{|} & \multicolumn{2}{l}{\texttt{<const>}} \\
    \texttt{<biop>}  & \texttt{::=} & \multicolumn{2}{l}{\texttt{+}} \\
                     &   \texttt{|} & \multicolumn{2}{l}{\texttt{-}} \\
                     &   \texttt{|} & \multicolumn{2}{l}{\texttt{*}} \\
                     &   \texttt{|} & \multicolumn{2}{l}{\texttt{/}} \\
                     &   \texttt{|} & \verb|^| & \textit{exponentiation}\\
    \texttt{<unop>}  & \texttt{::=} & \texttt{ln} & \textit{natural logarithm}\\
                     &   \texttt{|} & \texttt{exp} & \textit{natural exponential ($e^x$)}\\
                     &   \texttt{|} & \texttt{-} & \textit{unary minus}\\
                     &   \texttt{|} & \texttt{abs} & \textit{absolute value}\\
    \texttt{<var>}   & \texttt{::=} & \multicolumn{2}{l}{\texttt{x1 | x2 | ...}}\\
    \texttt{<const>} & \texttt{::=} & \multicolumn{2}{l}{\texttt{-1 | 1}}\\
\end{tabular}

\noindent
where \texttt{x1}, \texttt{x2}, etc. are the feature variables of the particular dataset.
\balancecolumns 
\end{document}